\documentclass[letterpaper]{article} 
\usepackage{aaai25}  
\usepackage{times}  
\usepackage{helvet}  
\usepackage{courier}  
\usepackage[hyphens]{url}  
\usepackage{graphicx} 
\usepackage{natbib}  
\usepackage{caption} 
\usepackage{algorithm}
\usepackage{algorithmic}

\usepackage{tabularx}
\usepackage{array}
\usepackage{makecell}

\usepackage{newfloat}
\usepackage{listings}
\DeclareCaptionStyle{ruled}{labelfont=normalfont,labelsep=colon,strut=off} 
\floatstyle{ruled}
\newfloat{listing}{tb}{lst}{}
\floatname{listing}{Listing}
\title{CasFT: Future Trend Modeling for Information Popularity Prediction with Dynamic Cues-Driven Diffusion Models}
\author{Xin Jing, Yichen Jing, Yuhuan Lu, Bangchao Deng, Xueqin Chen, and Dingqi Yang}
\usepackage{bibentry}
\usepackage{booktabs}
\usepackage{color,xcolor}
\usepackage{amssymb}
\usepackage{amsmath}
\graphicspath{{./pic/}}
\usepackage{multirow}
\usepackage{enumitem}
\usepackage{color}

\begin{document}

\maketitle

\begin{abstract}
The rapid spread of diverse information on online social platforms has prompted both academia and industry to realize the importance of predicting content popularity, which could benefit a wide range of applications, such as recommendation systems and strategic decision-making. 
Recent works mainly focused on extracting spatiotemporal patterns inherent in the information diffusion process within a given observation period so as to predict its popularity over a future period of time. However, these works often overlook the future popularity trend, as future popularity could either increase exponentially or stagnate, introducing uncertainties to the prediction performance. Additionally, how to transfer the preceding-term dynamics learned from the observed diffusion process into future-term trends remains an unexplored challenge.
Against this background, we propose CasFT, which leverages observed information \underline{Cas}cades and dynamic cues extracted via neural ODEs as conditions to guide the generation of \underline{F}uture popularity-increasing \underline{T}rends through a diffusion model. These generated trends are then combined with the spatiotemporal patterns in the observed information cascade to make the final popularity prediction.
Extensive experiments conducted on three real-world datasets demonstrate that CasFT significantly improves the prediction accuracy, compared to state-of-the-art approaches, yielding 2.2\%-19.3\% improvement across different datasets.

\end{abstract}


%
\section{Introduction}
\label{Intro}

The advent of the digital age has led to the emergence of various online social platforms (OSNs), such as Twitter and Facebook, revolutionizing the way of sharing information among people. Users on OSNs can freely post and share their interests, which can then be followed or re-posted by others, resulting in the widespread dissemination of content in the form of information cascades~\cite{cheng2014can}.
In such an era of information explosion, understanding future trends in information --- predicting its future popularity --- can aid social management~\cite{shen2014modeling} and benefit various applications, including recommendation systems~\cite{wang2021dydiff}, fake news detection~\cite{lazer2018science}, and personalized user experiences~\cite{li2014efficient}.
Specifically, this prediction task, known as information popularity prediction, aims at forecasting the future increase in the popularity of given content, such as predicting the number of retweets that may occur within a specified period in the future for a tweet.



\begin{figure}[t]\small
	\centering
	\includegraphics[width=0.48\textwidth]{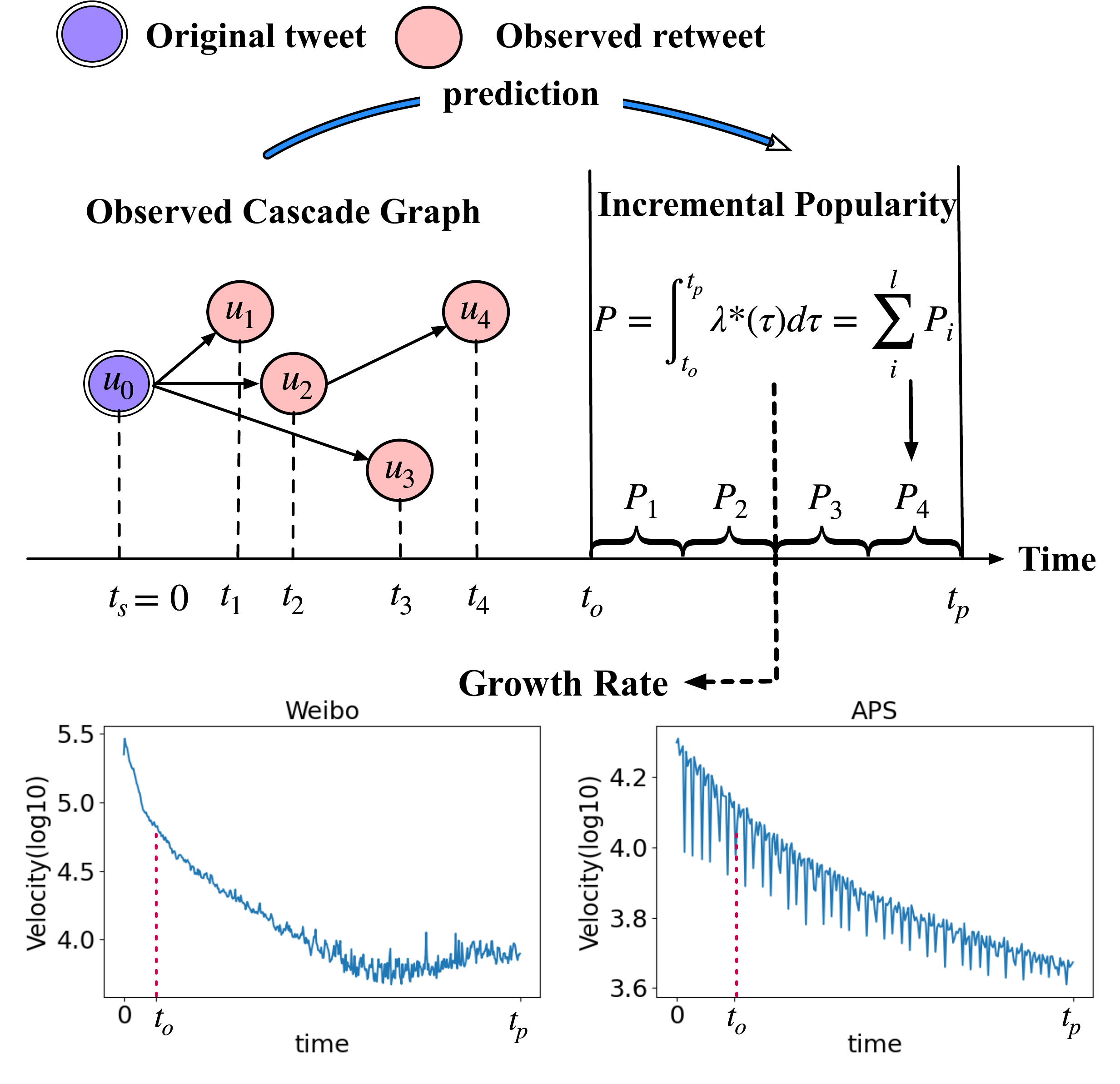}
	\caption{A toy example of information popularity prediction problem (top) and the variations of average growth rates after observation time in Weibo and APS datasets, respectively (bottom), where $t_0$ represents the observation time and $t_p$ represents the prediction time.}
    \vspace{-1em}
	\label{intro_pop}
\end{figure}

State-of-the-art approaches mainly focus on capturing spatiotemporal patterns from observed cascades by adopting various deep learning techniques, such as Recurrent Neural Networks (RNNs)~\cite{lu2023continuous}, attention mechanism~\cite{yu2022transformer} or Graph Neural Networks (GNNs)~\cite{wang2022casseqgcn,chen2019cascn}. However, existing approaches predominantly focus on modeling cascades before an observation time, neglecting their evolution of trends between the observation time and the prediction time, which is vital for popularity prediction. For instance, Figure \ref{intro_pop} illustrates the average popularity growth rates between observation ($t_o$) and prediction ($t_p$) times in Twitter and APS datasets, where the growth rate represents the increase of popularity per unit of time. Notably, during this period, the variation in growth rates fluctuates significantly across both two datasets, highlighting different cascade incremental patterns, which contributes to the uncertainty in depicting future diffusion trends. Subsequently, achieving robust popularity prediction necessitates accommodating such uncertainties by accounting for cascade evolution patterns after the observation period. Nevertheless, the actual dynamic changes in cascades during this period are invisible when performing real-time popularity prediction.

Against this background, we resort to generative models to simulate the cascade evolving patterns between the observation and prediction times for boosting popularity prediction performance. However, there exist two difficulties in characterizing the unique features in the patterns: 1) the incremental popularity equals the integral of the growth rate, which is highly fluctuated as evidenced by Figure \ref{intro_pop}; and 2) information diffusion is a complex propagation process that is susceptible to external factors, resulting in varying patterns of uncertainties that need to accommodate.

To address these challenges, we propose CasFT, a novel information popularity prediction technique that aims to capture the evolving patterns of information \underline{Cas}cades over time, in particular, the \underline{F}uture propagation \underline{T}rend. CasFT leverages neural Ordinary Differential Equations (ODEs)~\cite{chen2018neural} to model the growth rate based on corresponding graph structure and sequential event information under observation, propagate the growth rate from the observation time to the prediction time, and calculates the cumulative popularity during several periods by the integration of the growth rate. We take the historical information diffusion representation and the future dynamic cumulative popularity as conditions, adopt diffusion models to generate the future trend of information popularity, and fuse the cascade representation with the generated cues for prediction. In summary, our contributions are as follows:

\begin{itemize}[leftmargin=*]
\item We revisit the existing cascade popularity prediction methods and identify the importance of modeling future trends of popularity after observation which contributes to the final prediction.

\item We propose CasFT, which is designed to capture the evolving dynamic patterns of both the information cascades and growth rate with neural ODEs. We leverage diffusion models to generate future trends of popularity and concatenate them with dynamic cascade representations together for prediction.

\item We conduct extensive experiments to evaluate CasFT on several real-world information cascade datasets. Results show that CasFT achieves superior performance compared to a sizeable collection of state-of-the-art baselines on the information popularity prediction task, yielding 2.2\%-19.3\% improvement over the best-performing baselines across different datasets. 

\end{itemize}

\section{Related Work}
\label{RelatedWork}
In the current literature, cascade popularity prediction techniques can be roughly classified into three categories.

\noindent \textbf{Feature-based approaches:}
Feature-based approaches extract observed features manually which include information about the content, such as tags~\cite{ma2013predicting}, topic features~\cite{Martin2016}, and graph structures~\cite{gao2014effective,wang2016cpb}, temporal features such as publication time~\cite{petrovic2011rt,wu2016unfolding}, observation time~\cite{cheng2014can,yang2011patterns}, first participation time~\cite{zaman2014bayesian}, etc. However, these works heavily rely on the quality of hand-crafted features and thus are dataset-specific and have limited generalizability. 


\noindent \textbf{Generative-based approaches:} These methods use probabilistic models to describe the propagation, including epidemic models~\cite{matsubara2012rise}, survival analysis~\cite{lee2012modeling}, and various stochastic point processes, e.g., Poisson process~\cite{bao2015modeling}, Hawkes process~\cite{zhao2015seismic}, ~\cite{zhang2022anytime}. However, the selection of underlying probabilistic models can lead to significant differences in performance.


\noindent \textbf{Deep-learning-based approaches:} The recent rapid development of deep learning has also spurred the emergence of numerous information cascade models based on deep neural networks. 
DeepHawkes~\cite{cao2017deephawkes} constructs an end-to-end network by Hawkes processes. 
DeepCas~\cite{li2017deepcas} is the first cascade graph representation learning method, capturing both structural and temporal information. 
VaCas~\cite{zhou2020continual} and CasFlow~\cite{xu2021casflow} propose a hierarchical graph learning method using variational autoencoders or normalizing flows to model uncertainty in cascade graphs, and a bidirectional GRU to capture temporal dynamics.
CasSeqGCN~\cite{wang2022casseqgcn} aggregates cascade node representations using a dynamic routing mechanism.
CTCP~\cite{lu2023continuous}integrates multiple cascades into a diffusion graph to further capture the correlation between cascades and users. However, existing methods only consider modeling cascades within observation. We argue that CasFT incorporates the future trends of popularity, thereby improving the prediction performance.

\section{Preliminaries}
\label{Preliminaries}
In this section, we present the definitions of key concepts in our work and give a brief introduction to the concepts of neural ODEs and diffusion models.

\subsection{Problem Definition}
Suppose that a Twitter user \(u_{0}\) posts a tweet \(I\) at time \(t_s=0\) (we set the start time to 0) and other users can interact with tweet \(I\) through various actions such as liking, retweeting, and commenting. Here we focus on the ``retweet'' behavior, while our work can also be generalized to other actions. Specifically, when a user \(u_{1}\) retweets \(I\) posted by user \(u_{0}\) at time \(t_{1}\), we can define a triplet \((u_{0}, u_{1}, t_{1})\) to present the diffusion of information \(I\) from user \(u_{0}\) to user \(u_{1}\).
Subsequently, a series of such triplets constitutes a retweet cascade \(\mathcal{C}\). Specifically, within a given observation time \(t_{o}\), the retweet cascade \(\mathcal{C}(t_o)\) can be defined as the set of all relevant triplets, i.e., \(\mathcal{C}(t_{o})=\left \{(u_{i_{1}},u_{i_{2}},t_{i}) \right \} _{i \in N}\), where \(N\) represents the total number of triplets involved in the retweet process during the observation period and $t_i\leq t_o$.

\noindent \textbf{Cascade Graph:}
Given a tweet \(I\) and an observation time \(t_{o}\), the cascade graph can be defined as \(G(t_o)=(\mathcal{V}_{c}, \mathcal{E}_{c})\), where \(\mathcal{V}_{c}\) represents the set of users involved in \(\mathcal{C}(t_o)\), and \(\mathcal{E}_{c}\) represents the set of edges, each denoting a retweet relationship between two users (nodes).

\noindent \textbf{Cascade Sequence:} Different from the cascade graph, the cascade sequence $S(t_o)$ for cascade $\mathcal{C}(t_o)$ is a collection of users, each arranged in chronological order based on their timestamps, i.e., $S(t_o) = \{u_0,u_1,\ldots,u_N\}$.

\noindent \textbf{Global Graph:}
Given all the retweet cascades under the observation time $t_o$, we define the global graph as $\mathcal{G} = \left\{G_I(t_o)|I\in\mathcal{I}\right\}$, where the edge in $\mathcal{E}_g$ represents the node relationship from cascading, such as the follower/followee relationship in the social network.

\noindent \textbf{Cascade Popularity Prediction:}
Given a cascade \(\mathcal{C}(t_o)\) and a duration \( t_{o} \), we predict its incremental popularity \( P\) over the time interval from \(t_{o}\) to \(t_{p}\) , where \(t_{p} \gg t_{o}\), and \(t_{p}\) is the prediction time. \( P\) represents the number of triplets occurring during the time period $(t_o, t_p)$.

\subsection{Neural Ordinary Differential Equations}
A neural Ordinary Differential Equation (ODE)~\cite{chen2018neural} describes the continuous-time evolution of variables. It represents a transformation of variables over time, where the initial state at time $t_0$, denoted as $\mathbf{h}(t_0)$, is integrated forward using an ODE to determine the transformed state at any subsequent time $t_i$.

\begin{equation}
\label{ODE-1}
\frac{d\mathbf{h}(t)}{dt} = f (\mathbf{h}(t), t; \theta)\quad \text{where} \quad \mathbf{h}(0) = \mathbf{h}_0
\end{equation}
\begin{equation}
\label{ODE-2}
\mathbf{h}(t_i) = \mathbf{h}(t_0) + \int_{t_0}^{t_i}\frac{d\mathbf{h}(t)}{dt} dt,
\end{equation}
where $f$ denotes a neural network, such as a feed-forward or convolutional network.

\subsection{Diffusion Models}
Diffusion models~\cite{ho2020denoising} are used to generate high-quality samples from complex data distributions through a bi-directional process.


The forward process of diffusion model is a Markov Chain that gradually adds Gaussian noise to $x^0$:
\begin{equation}
\label{diff_forward1}
q(x^{1:T}|x^{0}) = \prod_{t=1}^{T}q(x^{t}|x^{t-1}),
\end{equation}

\begin{equation}
\label{diff_forward2}
q(x^{t}|x^{t-1})= \mathcal{N}(x^{t};\sqrt{1-\beta _{t}}x^{t-1},\beta_{t}I ),
\end{equation}
where $T$ is the timesteps of adding noise and $\beta_{t}\in (0,1)$ is the variance schedule.


For the denoising process, the goal is to learn a conditional probability distribution \(p_{\theta}(x^{t-1}|x^{t} ) \) that reverses the diffusion process:
\begin{equation}
\label{diff_denoising}
p_{\theta}(x^{t-1}|x^{t})=\mathcal{N}(x^{t-1};\mu_{\theta}(x^{t},t), \sigma_{\theta }(x^{t},t)I ), 
\end{equation}
where the parameter $\theta$ can be optimized by minimizing the negative log-likelihood via an Evidence Lower Bound (ELBO):

\begin{equation}
\label{theta_optimized}
\min_{\theta }{\mathbb{E}_{q(x^0)}} \le \min_{\theta } {\mathbb{E}_{q(x^0:K)}} L 
\end{equation}

\begin{equation}
\label{theta_optimized2}
L = -\log{p(x^K)}-\sum_{k=1}^{K}\log{\frac{p_{\theta }(x^{k-1}|x^k)}{q(x^k|x^{k-1})} } 
\end{equation}





	
\section{Our Method: CasFT}
\label{Method}
\begin{figure*}[htb]
	\centering
	\includegraphics[width=\textwidth]{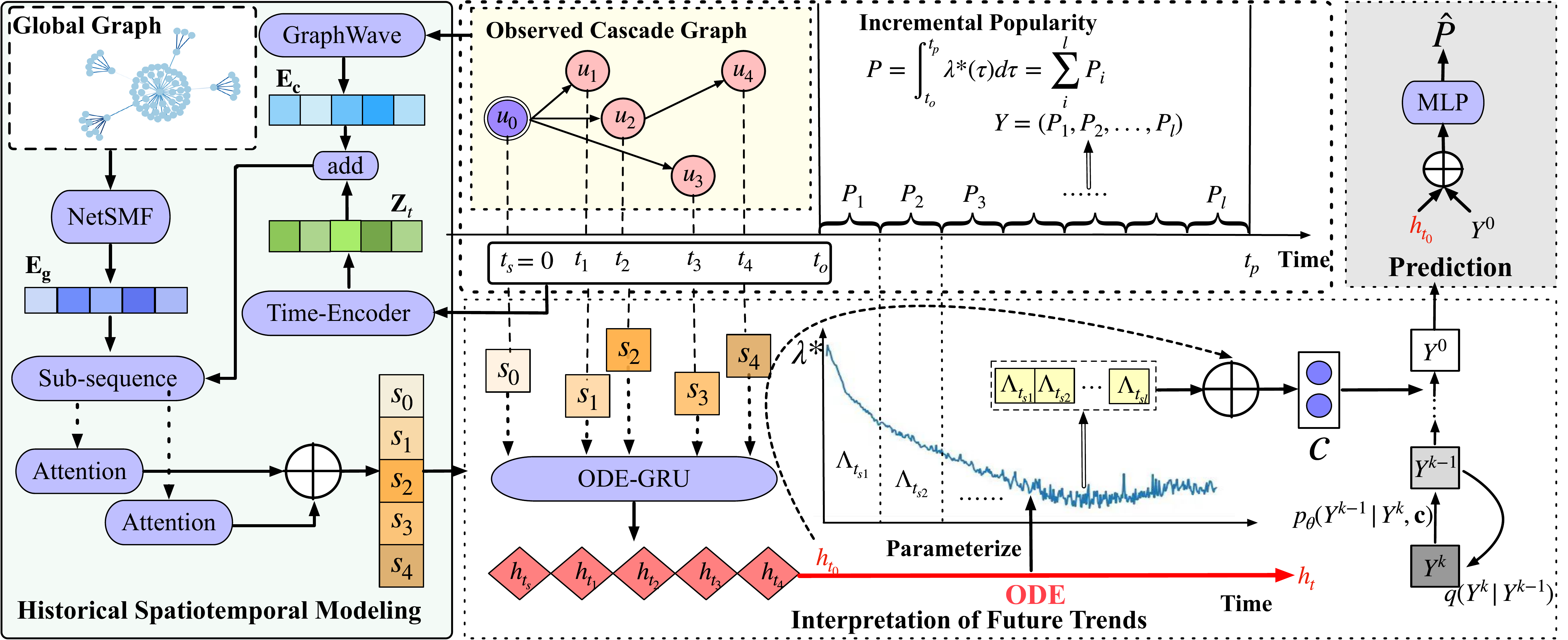}
	\caption{An overview of our proposed model CasFT.}
        \vspace{-1em}
	\label{CasFT}
\end{figure*}

Our proposed CasFT can be summarized as a three-step model: the first step involves extracting spatiotemporal patterns from the observed cascade, the second step simulates the future trends of popularity increment, and the third step makes the final prediction. The overall framework is shown in Fig.~\ref{CasFT}. We now provide a detailed discussion of its main components. 
\subsection{Historical Spatiotemporal Modeling}
Given an observed cascade $\mathcal{C}(t_0)$, CasFT first extracts the spatiotemporal features from it. Specifically, this step involves two components: (1) spatial structural pattern extraction from the observed cascade graph $G(t_0)$ and the global graph $\mathcal{G}$, and (2) temporal pattern extraction based on the cascade sequence. The details are outlined as follows.
\subsubsection{Spatial Structural Pattern Extraction:}
In light of the distinct characteristics between the cascade and global graphs, we employ two different graph embedding methods to extract unique structural patterns from each, respectively. For the cascade graph, we use GraphWave~\cite{donnat2018learning} to capture local structural information, as referenced in previous studies~\cite{zhou2020variational,xu2021casflow}. For the large-scale global graph, we employ NetSMF~\cite{qiu2019netsmf}, which is designed for large-scale networks and can quickly learn interactions between all users, efficiently embedding these global structural patterns into node embeddings. Specifically, given the observed cascade graph $G(t_o)$ and the global graph $\mathcal{G}$, we calculate the local-view and global-view representations of users as follows:
\begin{equation}
E_c = \text{GraphWave}(G(t_o)) ,
\end{equation}
\begin{equation}
E_g = \text{NetSMF}(\mathcal{G}) ,
\end{equation}
where $E_c$ and $E_g$ represent local-view and global-view representations for users in $G(t_o)$ and $\mathcal{G}$, respectively. For a specific user $u_i$, we retrieve its representations from these two views according to its index, denoted as $E_c(u_i)$ and $E_g(u_i)$. 




\subsubsection{Temporal Pattern Extraction:}
In this phase, we split the cascade sequence $S(t_o)$ into multiple sub-sequences in chronological order, represented as $\mathcal{S} =\{S_j\}_{j\in N} = \{(u_0),(u_0,u_1),\ldots,(u_0,u_1,...,u_N)\}$, and employ the self-attention mechanism~\cite{vaswani2017attention} to each sub-sequence to capture its internal interdependencies, thereby accounting for the long-term history information. Firstly, for each timestamp $t_i$, we define a temporal encoding procedure by trigonometric functions:
\begin{equation}
\label{t-encoding}
[\mathbf{z}(t_i)]_j = \left\{
\begin{aligned}
\cos(t_i/10000^{\frac{j-1}{B}}), \text{if j is odd,} \\
\sin(t_i/10000^{\frac{j}{B}}), \text{ if j is even.}
\end{aligned}
\right.
\end{equation}
We deterministically compute $\mathbf{z}(t_i)\in\mathbb{R}^B$, where $B$ is the dimension of encoding and get $\mathbf{Z}_t$ which is the set of $\mathbf{z}(t_i)$. The initial embedding of each sub-sequence $S_j$ is specified by 
\begin{equation}
\label{event-add}
\mathbf{x}(i) = \mathbf{z}(t_i) + E_c(u_i), 
\end{equation}
\begin{equation}
\label{embedding-sequence}
E_{c}^j=\{\mathbf{x}(0),\mathbf{x}(1),...,\mathbf{x}(i)\},
\end{equation}
\begin{equation}
\label{embedding-sequence2}
E_{g}^j=\{\ E_g(u_0), E_g(u_1),...,E_g(u_i)\},
\end{equation}
where $u_i \in S_j$, and $t_i$ is its associated time stamp.
We then encode $E_{c}^j$ and $E_{g}^j$ via two independent self-attention layers. Specifically, the scaled dot-product attention~\cite{vaswani2017attention} is defined as:
\begin{equation}
\label{Softmax}
\mathbf{s} = \text{Attention}(Q,K,V) = \text{Softmax}(\frac{QK^T}{\sqrt{d}})V,
\end{equation}
where $Q, K, V$ represent queries, keys, and values. In our case, the self-attention operation takes $E_{c}^j$ or $E_{g}^j$ as input and converts them into different groups of $Q$, $K$, and $V$ through separate linear transformations. 
By applying the above self-attention operation to all sub-sequences, we obtain sets of hidden representations $\mathbf{s}_{c} = \{\mathbf{s}_{c}^0,\mathbf{s}_{c}^1,\ldots,\mathbf{s}_{c}^N\}$ and $\mathbf{s}_{g} = \{\mathbf{s}_{g}^0,\mathbf{s}_{g}^1,\ldots,\mathbf{s}_{g}^N\}$ for the local-view and global-view, respectively. Finally, we concatenate these two views of hidden representations at the sub-sequence level to form the final historical spatiotemporal features of $\mathcal{S}$, denoted by $\{\mathbf{s}_{0},\mathbf{s}_{1},\ldots,\mathbf{s}_{N}\}$.

\subsection{Interpretation of Future Trends}
As we discussed in the Introduction section, the real dynamics of cascade evolution between the observation and prediction periods are invisible and difficult to depict due to their fluctuations. 
To address this challenge, we designed a dynamic cues-driven diffusion model to simulate future trends based solely on the observed cascades.
Specifically, this model consists of two components: (1) a dynamic cues learning module and (2) a diffusion model-based future trends simulation module. To acquire the dynamic cues, we are inspired by ODEs, which are used to model quantity changes over time in dynamic systems. We introduce a neural ODEs to model the continuous-time growth rate during the observation period, capturing its evolving tendency. Afterward, the future growth rate between the observation and prediction periods can be autoregressively predicted using this neural ODEs. Finally, the dynamic cues of popularity increment are derived through an integration operation based on these predicted growth rates. However, one limitation of ODEs is that they are constrained by the initial state of the dynamic system, leading to popularity predictions based on the estimated growth rate that are biased and lack realistic variability. Accordingly, we further divide the estimated period into several time spans and employ diffusion models to generate incremental popularity in each time span, conditioned on the growth rate.

\subsubsection{Dynamic Cues Learning with Neural ODEs:}
We leverage neural ODEs to model the continuous-time growth rate dynamics with a vector representation $\mathbf{h}_{t_i}$ at every timestamp $t_i$.
Here we parameterize the growth rate using hidden state dynamics:
\begin{equation}
\label{growth_rate}
\lambda^*(t) = f_1({\mathbf{h}}_{t}),
\end{equation}
where $f$ is a linear neural network with the softplus function to ensure the output is positive. 
As for the continuous-time hidden state $h_{t_i}$, we use a multi-layer fully connected neural network $f_2$ to model the continuous change in the form of an ODE. When a new retweeting action occurs at time $t_i$, we use a GRU function $g$ to model instantaneous changes triggered by a newly observed event:
\begin{equation}
\label{ODE-t}
\frac{d{\mathbf{h}}_{t_i}}{dt} = f_2 (t, {\mathbf{h}}_{t_{i-1}} ) ,
\end{equation}
\begin{equation}
\label{ode-solver}
{\mathbf{h}}_{t_i}^{'} = \text{ODESolve}(f_2,\mathbf{h}_{t_{i-1}},(t_{i-1},t_i)),
\end{equation}
\begin{equation}
{\mathbf{h}}_{t_i} = g({\mathbf{h}}_{t_i}^{'},\mathbf{s}_i),
\end{equation}
and the popularity at time $t_i$ can be calculated as:
\begin{equation}
\label{integration}
P_{t_i} = \int_0^{t_i}{\lambda}^*(\tau)d\tau = \Lambda_{t_{i}}.
\end{equation}
Then we can get the hidden state, growth rate, and cumulative popularity at an arbitrary time by the above equations, even after observation time. 



\subsubsection{Future Trends Simulation with Diffusion Model:}
We first divide the time between the observation and prediction time into $l$ uniform time intervals, denoted as $\{(t_o, t_{s1}),..., (t_{si-1}, t_{si}),...,(t_{sl-1},t_{sl})\}$, where $i\in l$. The start time is $t_o$ and the end time $t_{sl}$ equals to $t_p$. We aim to model the incremental popularity in each time interval and define $Y=(P_1,P_2,...,P_l)$ as the sequential increase process of popularity. $y_i$ represents the popularity during the time interval $(t_{i-1}, t_{i})$. 

We compute the hidden state $\mathbf{h}_{t_o}$ at the exact observation time, and the dynamic cues $\{\Lambda_{t_{s1}},\ldots, \Lambda_{t_{sl}}\}$ by using Eq.~\eqref{ODE-t} and~\eqref{integration}, respectively. We then concatenate them together to form a new guide vector $\mathbf{c}$.
Conditioned on $\mathbf{c}$, we utilize the diffusion model to approximate the real distribution of the incremental popularity sequence $Y$. Specifically, we characterize the diffusion operation of $Y$ as a Markov process $(Y^0, Y^1, ..., Y^K)$ to model the potential changes of incremental popularity over the prediction window, with $T$ denoting the number of diffusion steps. In the forward process, Gaussian noise is added to $Y^0$ step by step until it is corrupted into pure Gaussian noise, which is formulated as follows:
\begin{equation}
\label{diff_forward_incremental}
q(Y^{k}|Y^{k-1})= \mathcal{N}(Y^{k};\sqrt{1-\beta _{k}}Y^{k-1},\beta_{k}I).
\end{equation}
During the reverse diffusion process, we iteratively reconstruct the incremental popularity sequence $Y^0$ for the final prediction. Specifically, the denoising process does not only depend on the representation obtained in the previous step but also on the significant cue $\mathbf{c}$ serving as a condition, which is formulated as follows:
\begin{equation}
\label{diff_backward_incremental}
p_{\theta}(Y^{0:K}|\mathbf{c}) := p(Y^K)\prod_{k=1}^{K} p_{\theta}(Y^{k-1}|Y^k,\mathbf{c}),
\end{equation}
In our implementation, we adopt a fully connected layer to parameterize the denoising process and use Denoising Diffusion Implicit Models (DDIM) ~\cite{song2020denoising} to generate segmented popularity sequence $Y$.

\subsection{Prediction \& Optimization}
\noindent\textbf{Prediction.} After obtaining both the hidden state $\textbf{h}_{t_o}$ at observation time $t_o$ and the generated segmented popularity sequence $Y^0$, we concatenate and feed them into an MLP to make the final cascade popularity prediction:
\begin{equation}
{ \hat P} = \text{Softplus}(\text{MLP}([Y^0, \textbf{h}_{t_o}]))
\end{equation}
where the softplus activation function is to ensure the predicted popularity is positive. 

\noindent\textbf{Optimization.} The training objective of CasFT can be decomposed into a regression loss and a generative loss. As for the regression part, the loss function is defined as:
\begin{equation}
\begin{split}
\mathcal{L}_1 = \frac{1}{M}\sum_{k=1}^{M} {({\log}_2(P+1) - {\log}_2( \hat{P}+1))}^2 ,
\end{split}
\end{equation}
where $P$ and $\hat{P}$ is the ground-truth and the predicted popularity, respectively. For the generative part, we minimize the following negative log-likelihood:
\begin{equation}
\begin{split}
\mathcal{L}_2 = - \sum_{k=1}^{M} \log{p_{\theta}(Y_k^{0}|c_k}).
\end{split}
\end{equation}
To sum up, the final loss is defined as:
\begin{equation}
\begin{split}
\mathcal{L} = \mathcal{L}_1 +  \gamma\mathcal{L}_2,
\end{split}
\end{equation}
where $\gamma$ is a hyperparameter used to adjust the trade-off between two losses.
	
\section{Experiments}
\label{experiments}
\begin{table}
\centering
\small
\begin{tabular}{c c c c}
\Xhline{1px}
Dataset & Twitter & APS & Weibo\\
\hline
Cascades & 86,764  & 207,685 & 119,313\\
Avg. popularity & 94  & 51 & 240\\
\hline
\multicolumn{4}{c}{ \emph{Number of cascades in two observation settings}} \\
Train(1d/0.5h/3y) & 7,308  & 18,511 & 21,463\\
Val(1d/0.5h/3y) & 1,566 & 3,967 & 4,599 \\
Test(1d/0.5h/3y) & 1,566  & 3,966 & 4,599\\
Train(2d/1h/5y) & 10,983  & 32,102 & 29,908\\
Val(2d/1h/5y) & 2,353  & 6,879 & 6,409\\
Test(2d/1h/5y) & 2,353  & 6,879 & 6,408\\
\hline
\Xhline{1px}

\end{tabular}
\caption{Statistics of the three datasets.}
\vspace{-1em}
\label{dataset}
\end{table}

\begin{table*}
    \centering    
    \small
    \begin{tabularx}{\textwidth}{@{} c| *{1}{>{\centering\arraybackslash}X} *{1}{>{\centering\arraybackslash}X|} *{1}{>{\centering\arraybackslash}X} *{1}{>{\centering\arraybackslash}X|} *{1}{>{\centering\arraybackslash}X} *{1}{>{\centering\arraybackslash}X|} *{1}{>{\centering\arraybackslash}X} *{1}{>{\centering\arraybackslash}X|} *{1}{>{\centering\arraybackslash}X} *{1}{>{\centering\arraybackslash}X|} *{1}{>{\centering\arraybackslash}X} *{1}{>{\centering\arraybackslash}X} *{1}{>{\centering\arraybackslash}X} @{}} 
    
    \hline
    \Xhline{1px}
    
    \multirow{3}{*}{ \textbf{Method} } &
    
    \multicolumn{4}{c|}{\textbf{Twitter}} &
    \multicolumn{4}{c|}{\textbf{APS}} &
    \multicolumn{4}{c}{\textbf{Weibo}} \\
    \cline{2-13}

~ &
    \multicolumn{2}{c|}{\textbf{1 day}} &
    \multicolumn{2}{c|}{\textbf{2 days}} &
    \multicolumn{2}{c|}{\textbf{3 years}} &
    \multicolumn{2}{c|}{\textbf{5 years}} &
    \multicolumn{2}{c|}{\textbf{0.5 hours}} &
    \multicolumn{2}{c}{\textbf{1 hour}} \\
\cline{2-13}
    
     & MSLE & MAPE & MSLE & MAPE & MSLE & MAPE & MSLE & MAPE & MSLE & MAPE & MSLE & MAPE \\
     \hline
    \hline
    Feature-based & 7.8268 & 0.7073 & 6.5154 & 0.6514 & 1.9881 & 0.3085 & 1.9696 & 0.3193 & 4.0788 & 0.4094 & 3.6380 & 0.4268 \\
    SEISMIC & 10.687 & 0.9689 & 8.1851 & 0.8147 & 2.0583 & 0.3013 & 2.3013 & 0.4320 & 5.0300 & 0.4819 & 4.0594 & 0.5003 \\
    DeepCas & 6.3297 & 0.6566 & 5.7146 & 0.6671 & 2.1051 & 0.2869 & 1.9260 & 0.3458 & 4.6460 & 0.3258 & 3.5532 & 0.3532 \\
    DeepHawkes & 5.9341 & 0.5017 & 4.8489 & 0.5189 & 1.9142 & 0.2823 & 1.8145 & 0.3368 & 2.8741 & 0.3041 & 2.7434 & 0.3346 \\
    CasCN & 5.8742 & 0.4894 & 4.7154 & 0.4974 & 1.8930 & 0.2763 & 1.7494 & 0.3208 & 2.7931 & 0.2940 & 2.6831 & 0.3255 \\
    VaCas & 5.5124 & 0.4796 & 4.2147 & 0.4871 & 1.7764 & 0.2697 & 1.6945 & 0.3012 & 2.5246 & 0.2847 & 2.3451 & 0.2997 \\
    CasFlow &  \underline{4.7799} & 0.4150 & 3.6888 & 0.4222 & \underline{1.4370} & \underline{0.2401} & \underline{1.3346} & \underline{0.2624} & \underline{2.3370} & \underline{0.2665} & \underline{2.2232} & \underline{0.2949} \\
    CTCP & 5.3991 & \underline{0.3757} & \underline{3.6016} & \underline{0.3773} & 1.7676 & 0.3054 & 1.3751 & 0.2908 & 2.5572 & 0.3056 & 2.2968 & 0.3010 \\
        \hline
    \textbf{CasFT} & \textbf{3.8546} & \textbf{0.3674} & \textbf{3.4496} & \textbf{0.3605} & \textbf{1.2468} & \textbf{0.2282} & \textbf{1.1748} & \textbf{0.2561} & \textbf{2.1728} & \textbf{0.2448} & \textbf{2.0655} & \textbf{0.2695} \\
    
    (Improve) & 19.36\%$\uparrow$ & 2.21\%$\uparrow$ & 4.22\%$\uparrow$ & 4.45\%$\uparrow$ & 13.24\%$\uparrow$ & 4.96\%$\uparrow$ & 11.97\%$\uparrow$ & 2.40\%$\uparrow$ & 7.02\%$\uparrow$ & 8.14\%$\uparrow$ & 10.07\%$\uparrow$ & 8.61\%$\uparrow$ \\
    
    \Xhline{1px}
    \hline
    \end{tabularx}
    \caption{Performance comparison between baselines and CasFT on three datasets across different observation times measured by MSLE, MAPE (lower is better).}
    \vspace{-1em}
    \label{results}
\end{table*}

\begin{table*}
    \centering    
    \small
    \begin{tabularx}{\textwidth}{@{} c| *{1}{>{\centering\arraybackslash}X} *{1}{>{\centering\arraybackslash}X|} *{1}{>{\centering\arraybackslash}X} *{1}{>{\centering\arraybackslash}X|} *{1}{>{\centering\arraybackslash}X} *{1}{>{\centering\arraybackslash}X|} *{1}{>{\centering\arraybackslash}X} *{1}{>{\centering\arraybackslash}X|} *{1}{>{\centering\arraybackslash}X} *{1}{>{\centering\arraybackslash}X|} *{1}{>{\centering\arraybackslash}X} *{1}{>{\centering\arraybackslash}X} *{1}{>{\centering\arraybackslash}X} @{}} 
    
    \hline
    \Xhline{1px}
    
    \multirow{3}{*}{ \textbf{Method} } &
    
    \multicolumn{4}{c|}{\textbf{Twitter}} &
    \multicolumn{4}{c|}{\textbf{APS}} &
    \multicolumn{4}{c}{\textbf{Weibo}} \\
    \cline{2-13}

~ &
    \multicolumn{2}{c|}{\textbf{1 day}} &
    \multicolumn{2}{c|}{\textbf{2 days}} &
    \multicolumn{2}{c|}{\textbf{3 years}} &
    \multicolumn{2}{c|}{\textbf{5 years}} &
    \multicolumn{2}{c|}{\textbf{0.5 hours}} &
    \multicolumn{2}{c}{\textbf{1 hour}} \\
\cline{2-13}
    
     & MSLE & MAPE & MSLE & MAPE & MSLE & MAPE & MSLE & MAPE & MSLE & MAPE & MSLE & MAPE \\
     \hline
    \hline
    CasFT-w/o FT & 4.8641 & 0.3964 & 4.1154 & 0.4531 & 1.4147 & 0.2504 & 1.3544 & 0.2805 & 2.3964 & 0.2745 & 2.3412 & 0.2933 \\
    CasFT-w/o ODE & 4.3115 & 0.3902 & 3.7468 & 0.4174 & 1.3241 & 0.2473 & 1.2366 & 0.2778 & 2.3054 & 0.2601 & 2.2247 & 0.2794 \\
    CasFT-w/o Diffusion & 4.2645 & 0.3887 & 3.6618 & 0.4037 & 1.3258 & 0.2466 & 1.2364 & 0.2776 & 2.3312 & 0.2631 & 2.2644 & 0.2819 \\
    \hline
    CasFT-FM & 4.5334 & 0.3875 & 4.0047 & 0.4329 & 1.3847 & 0.2495 & 1.3017 & 0.2798 & 2.3645 & 0.2681 & 2.3015 & 0.2884 \\
    \hline
    \textbf{CasFT} & \textbf{3.8546} & \textbf{0.3674} & \textbf{3.4496} & \textbf{0.3605} & \textbf{1.2468} & \textbf{0.2282} & \textbf{1.1748} & \textbf{0.2561} & \textbf{2.1728} & \textbf{0.2448} & \textbf{2.0655} & \textbf{0.2695} \\
    \Xhline{1px}
    \hline
    \end{tabularx}
    \caption{Performance comparison between CasFT and CasFT-variants on three datasets under two observation times measured by MSLE, MAPE (lower is better).}
    \vspace{-1em}
    \label{ablation}
\end{table*}

In this section, we first present our benchmark datasets and then evaluate our model against state-of-the-art baselines in information cascade popularity prediction task to answer the following questions:

{\textbf{RQ1:}} Compared to state-of-the-art baselines, can our approach achieve a more accurate prediction of cascade popularity?

{\textbf{RQ2:}} What are the benefits of employing neural ODEs to model the growth rate? How much does it contribute to performance improvement?

{\textbf{RQ3:}} Why do we need diffusion models for enhancing the modeling of future trend dynamics? In contrast, how does the utilization of simpler models capture this characteristic? 

{\textbf{RQ4:}} What is the impact of the hyperparameters, including the number of segmented periods, the choice of ODESolver, the hidden dimension, and the diffusion steps?

\subsection{Datasets and Preprocessing}

We conducted experiments on three real-world datasets, including Twitter, APS~\cite{shen2014modeling}, and Sina Weibo~\cite{cao2017deephawkes}. The details of these datasets are as follows:

\begin{itemize}
    \item \textbf{\textit{Twitter}}: We collected tweets posted between March 1 and April 15, 2022, constructing the cascades based on the hashtag, where original tweets are posted between March 1 and March 31.
    \item \textit{American Physical Society} (\textbf{\textit{APS}}): This dataset includes papers published in APS journals between 1893 and 1997, where each paper and its citations form a citation cascade. 
    \item \textit{Sina \textbf{Weibo}}: Weibo is the largest Chinese microblogging platform, where each original microblog post and subsequent reposts can form a repost cascade. 
\end{itemize}

Following previous works~\cite{xu2021casflow, lu2023continuous}, we set observation time as 1 day and 2 days for the Twitter dataset, and 3 years and 5 years for the APS dataset, for Weibo, the observation times are set to 0.5 hours and 1 hour. Furthermore, prediction time is set to 15 days for Twitter, 20 years for APS, and 24 hours for Weibo. In addition, we filter out cascades with fewer than 10 participants during the observation periods. For all datasets, $70\%$ of the data is used for training, 15 $\%$ for validation, and 15$\%$ for testing. Detailed information about the three datasets is provided in Table~\ref{dataset}.

\subsection{Baselines}

We compare CasFT against a sizeable collection of state-of-the-art baselines: \textbf{Feature-based} methods extract key features (cascade size, temporal intervals, etc.) and use an MLP model for prediction.
\textbf{SEISMIC}~\cite{zhao2015seismic} designs a statistical model based on the theory of self-excited point processes. \textbf{DeepCas}~\cite{li2017deepcas} represents cascades as random walk paths and uses a bi-directional GRU with attention for effective modeling and prediction.
\textbf{DeepHawkes}~\cite{cao2017deephawkes} integrates the Hawkes process and deep learning, focusing on user impact, self-excitation, and time decay for cascade modeling.
\textbf{CasCN}~\cite{chen2019cascn} adopts a novel multi-directional/dynamic GNN.
\textbf{Vacas}~\cite{zhou2020variational} devises a hierarchical graph learning method and considers the uncertainty.
\textbf{CasFlow}~\cite{xu2021casflow} mainly considers the effects of local and global graphs.
\textbf{CTCP}~\cite{lu2023continuous} updates hidden states in real-time as diffusion events occur.

\subsection{Evaluation Metrics}
Two commonly used metrics, mean squared logarithmic error (MSLE) and mean absolute percentage error (MAPE), are employed to evaluate the performance of models:

\begin{equation}
\vspace{-1em}
MSLE = \frac{1}{M}\sum_{k=1}^{M}(\log_{2}{( P+1)}- (\log_{2}{( \hat{P}+1)})^{2} ,
\end{equation}

\begin{equation}
MAPE = \frac{1}{M}\sum_{k=1}^{M}\frac{\left | \log_{2}{( P+2)} -\log_{2}{( \hat{P} +2)} \right |  }{\log_{2}{( P+2)}} ,
\end{equation}


\noindent where \( P\) represents the true popularity, \( \hat{P} \) denotes the predicted popularity, and \(M\) represents the number of cascades. The operations '+ 1' and '+ 2' are used for scaling to avoid potential zero values in the denominator or logarithmic operations.

\subsection{Performance Comparison (RQ1)}
Table~\ref{results} shows the overall performance of the three datasets. We observe that our method CasFT significantly outperforms all baselines on MSLE and MAPE. For example, compared to the best-performing baselines, our CasFT achieves a 4.2\%-19.3\% improvement of MSLE, and a 2.2\%-8.6\% improvement of MAPE, on the three datasets with two different observation times. We see that CasFT demonstrates notable enhancement, especially on the Twitter dataset. This can be attributed to the fact that cascades on Twitter often lack intricate multi-level forwarding information, resulting in a relatively simpler structure of the cascade graph compared to the Weibo and APS datasets while previous research predominantly concentrates on the evolution of dynamic graphs, underscoring the versatility of our proposed method, CasFT, in effectively handling various types of cascade graphs.

\subsection{Ablation Study (RQ2 \& RQ3)}
To answer RQ2 and RQ3, we have conducted a series of experiments where we introduced different variants of our CasFT model.
We conduct the ablation study to investigate the contribution of each component and develop four variants: 1) \textbf{CasFT-w/o FT}. We remove the future trend modeling module and only use the spatiotemporal features $\mathcal{S}$ for prediction; 2) \textbf{CasFT-w/o ODE}. We take the spatiotemporal features $\mathcal{S}$ as the condition of the later diffusion models, without parameterizing the growth rate; 3) \textbf{CasFT-w/o Diffusion}. We remove the diffusion block and directly input the condition $c$ into an MLP; and 4) \textbf{CasFT-FM}. We just use an MLP to replace our future trend modules, predict both the segmented popularity and the incremental popularity during $(t_o, t_p)$, and also take the predicted segmented popularity sequence as a significant cue.


The results and comparison of these variants are shown in Table~\ref{ablation}. The comparison of CasFT over CasFT-w/o FT validates the necessity of modeling the future trend of cascade with an improvement of up to 20.75\%. Through a comparative analysis of CasFT-w/o FT, CasFT-w/o ODE, and CasFT-w/o Diffusion, it becomes evident that modeling the growth rate and segmented popularity generation both facilitate prediction accuracy, achieving improvements of 20.75\%, 13.63\%, and 10.70\% respectively. We also designed a variant CasFT-FM which outperforms CasFT-w/o FT but is worse than CasFT, showing the usefulness of our proposed future trend block.

\begin{figure}
	\begin{minipage}[t]{0.32\linewidth}
		\centerline{\includegraphics[width=\textwidth]{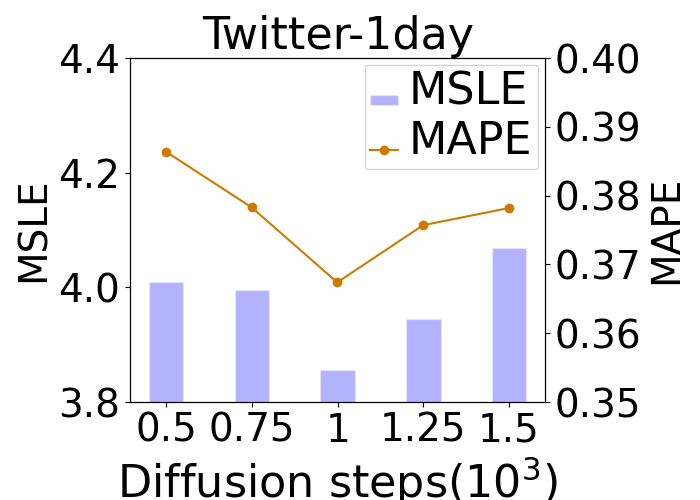}}
	\end{minipage}
	\begin{minipage}[t]{0.32\linewidth}
		\centerline{\includegraphics[width=\textwidth]{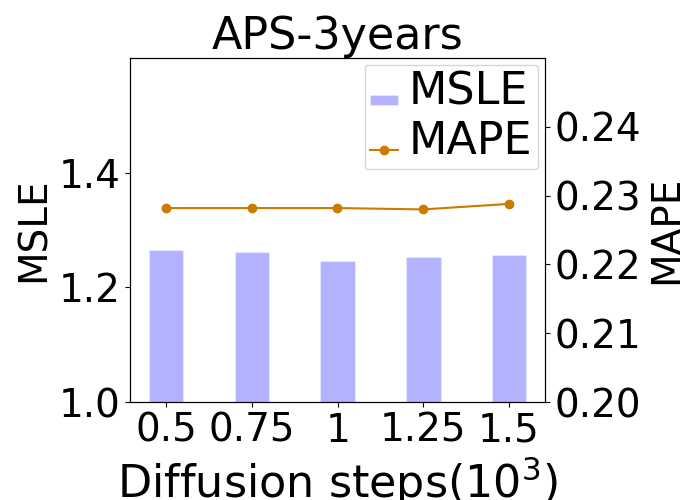}}
	\end{minipage}
	\begin{minipage}[t]{0.32\linewidth}
		\centerline{\includegraphics[width=\textwidth]{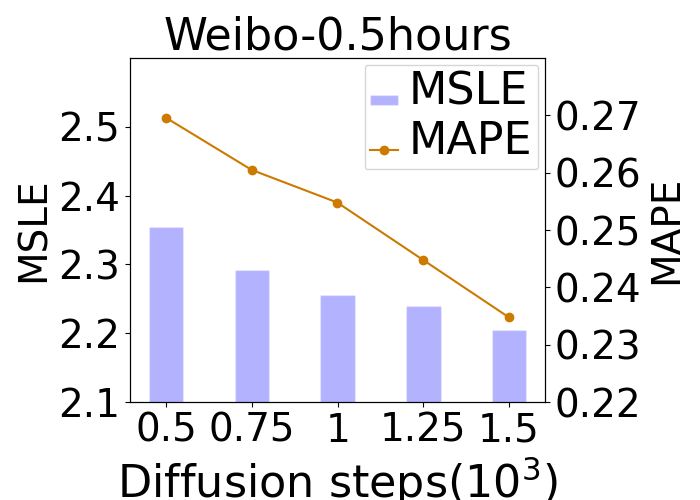}}
	\end{minipage}
	\caption{Impact of the diffusion steps.}
        \vspace{-1em}
        \label{Hyperparameters1}
\end{figure}

\begin{figure}
	\begin{minipage}[t]{0.32\linewidth}
		\centerline{\includegraphics[width=\textwidth]{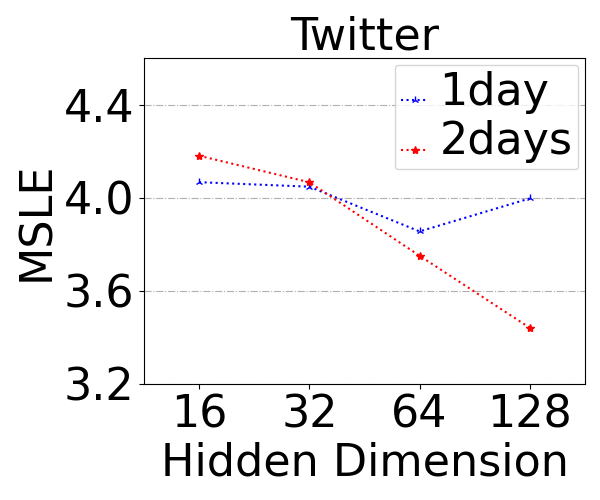}}
	\end{minipage}
	\begin{minipage}[t]{0.32\linewidth}
		\centerline{\includegraphics[width=\textwidth]{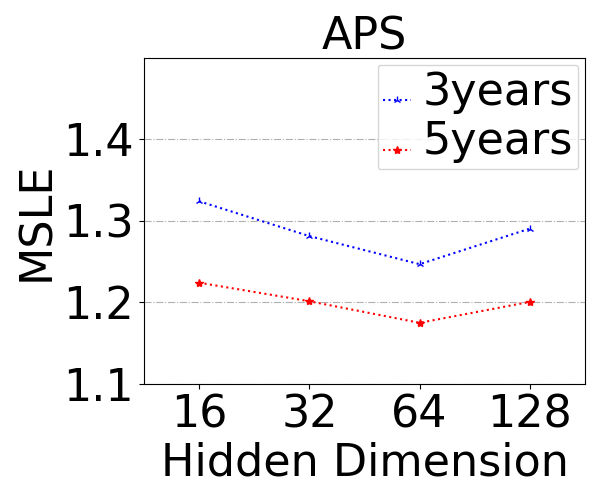}}
	\end{minipage}
	\begin{minipage}[t]{0.32\linewidth}
		\centerline{\includegraphics[width=\textwidth]{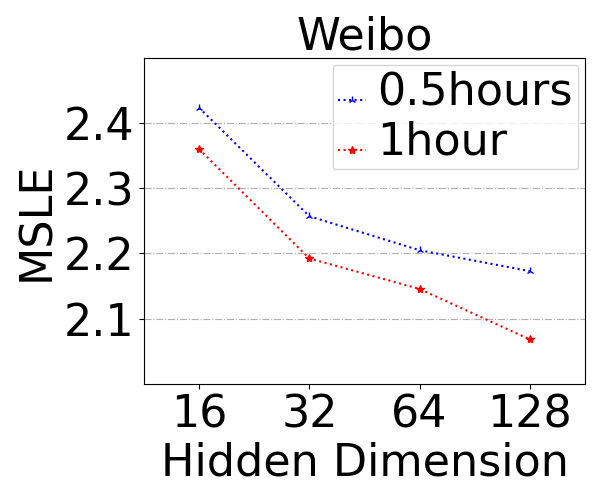}}
	\end{minipage}
	\caption{Impact of the hidden dimension.}
        \vspace{-1em}
        \label{Hyperparameters2}
\end{figure}

\begin{figure}
	\begin{minipage}[t]{0.32\linewidth}
		\centerline{\includegraphics[width=\textwidth]{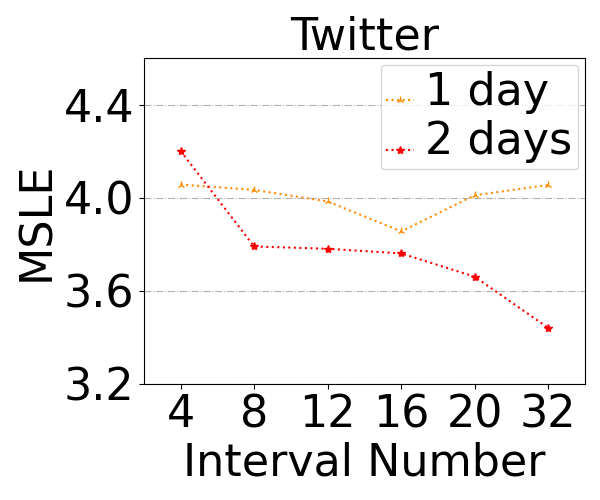}}
	\end{minipage}
	\begin{minipage}[t]{0.32\linewidth}
		\centerline{\includegraphics[width=\textwidth]{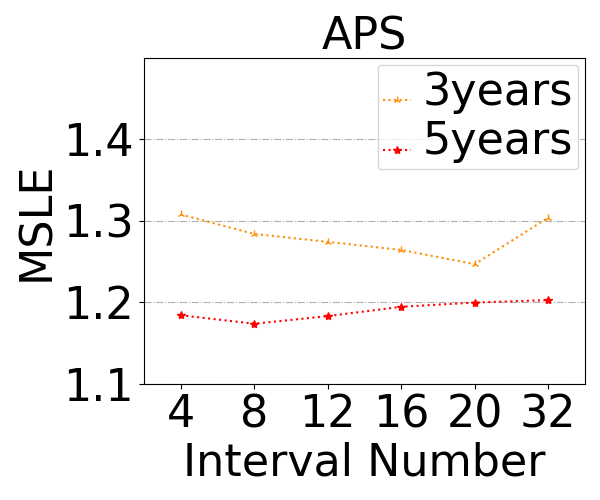}}
	\end{minipage}
	\begin{minipage}[t]{0.32\linewidth}
		\centerline{\includegraphics[width=\textwidth]{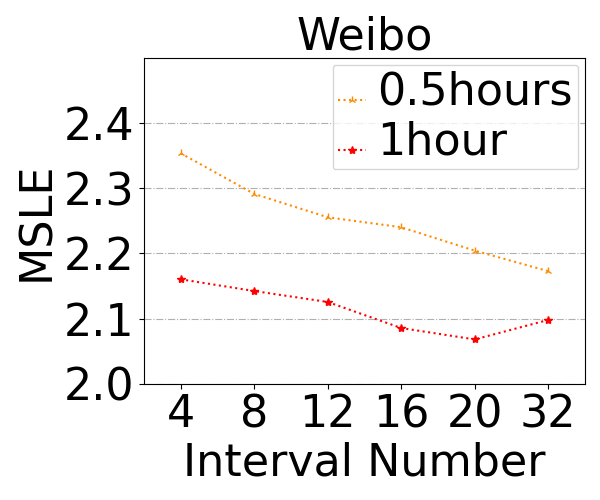}}
	\end{minipage}
	\caption{Impact of the interval number.}
        \vspace{-1em}
        \label{Hyperparameters3}
\end{figure}

\begin{table}
\small
\centering
\setlength{\tabcolsep}{1mm}{
\begin{tabular}{l | c  c | c c | c c }
\Xhline{1px}
\multirow{2}{*}{ \textbf{ODE\_Solver} } & \multicolumn{2}{c|}{Twitter (1 day)} & \multicolumn{2}{c|}{APS (3 years)} & \multicolumn{2}{c}{Weibo (0.5 hours)} \\
\cline{2-7}
 & MSLE & MAPE & MSLE & MAPE & MSLE & MAPE\\
\hline

bosh3 & 4.0933 & 0.3764  & 1.2701 & 0.2292 & 2.1792 & 0.2509 \\
adaptive\_heun & 4.2215 & 0.3995 & 1.2702 & 0.2291 & 2.1786 & 0.2578  \\
euler & 4.2487 & 0.3989 & 1.2837 & 0.2304 & 2.2188 & 0.2598 \\
rk4 & 4.2115 & 0.3720 & 1.2644 & 0.2263 & 2.1780 & 0.2537 \\
implicit\_adams & 4.1503 & 0.3722 & 1.2644 & 0.2263  & 2.1780 & 0.2537 \\
midpoint & 4.0760 & 0.3697 & 1.2622 & 0.2280 & 2.1873 & 0.2457  \\
dopri5 & \textbf{3.8546} & \textbf{0.3679}& \textbf{1.2468} & \textbf{0.2282} & \textbf{2.1728} & \textbf{0.2448} \\
\Xhline{1px}
\end{tabular} }
\caption{Impact of different types of ODE\_Solver on the performance of CasFT across the three datasets.}
\vspace{-1em}
\label{hyper_para_odesolver}
\end{table}

\subsection{Hyperparameters (RQ4)}



We investigate the influence of hyperparameters including the hidden dimension of ${h}_{t}$, the choice of ODESolvers, the number of the time interval $l$, and the diffusion steps. Firstly, the diffusion steps serve as a crucial parameter for diffusion models. To assess the influence of diffusion steps, we train CasFT using varying steps, ranging from 500 to 1500. The results are shown in Figure~\ref{Hyperparameters1} and we find that the number of diffusion steps has an impact on the Twitter dataset while the MAPE in APS and Weibo datasets tend to be stable. Additionally, among the ODESolver options shown in Table~\ref{hyper_para_odesolver}, euler yields the worst performance because the error of euler's method usually decreases as the step size decreases, while dopri5 demonstrates relatively superior predictive capabilities and is used across all of our experiments.

Moreover, the prognostication of CasFT relies on the dynamic hidden state $h_t$ and the generated segmented popularity sequence $Y^0$, prompting an investigation into the influence of the dimensionality of $h_t$ and the interval number $l$ of $Y^0$ on the performance, shown in Figure~\ref{Hyperparameters2} and Figure~\ref{Hyperparameters3}. It is observed that both the dimension of the hidden state and the number of intervals affect the ultimate forecasting outcomes.

\section{Conclusion}
\label{conclusion}

In this work, we propose CasFT, which leverages observed information \underline{Cas}cades and dynamic cues modeled via neural ODEs as conditions to guide the generation of \underline{F}uture popularity-increasing \underline{T}rends through a diffusion model. 
The generated trends are integrated with the spatiotemporal patterns present in the observed information cascades to enhance the accuracy of popularity predictions.
Experiments on three real-world information cascade datasets demonstrate the superior performance of CasFT compared to a sizeable collection of state-of-the-art baselines. CasFT significantly outperforms all the baselines, with 2.2\%-19.3\% improvement over the best-performing baseline methods across various datasets. 


\bibliography{0-CasDiff}

\end{document}